\documentclass[runningheads]{llncs}
\usepackage{float}
\usepackage[T1]{fontenc}
\usepackage{appendix}
\usepackage{multirow}
\usepackage[table,xcdraw]{xcolor}
\usepackage[normalem]{ulem}
\useunder{\uline}{\ul}{}
\usepackage{graphicx}
\usepackage{todonotes}
\usepackage{blindtext}
\usepackage{tcolorbox}
\usepackage{listings}
\usepackage{booktabs} 
\usepackage{amsmath} 
\usepackage{subcaption}
\usepackage{lscape}
\usepackage{color}
\begin{document}
\title{AgentEval: Generative Agents as Reliable Proxies for Human Evaluation of AI-Generated Content}
%
%
\author{Thanh Vu\inst{1}\orcidID{0009-0000-4152-2196} \and
Richi Nayak\inst{1}\orcidID{0000-0002-9954-0159
} \and
Thiru Balasubramaniam\inst{1}\orcidID{0000-0002-8821-6003} }
 
\authorrunning{Thanh Vu et al.}
\institute{Centre for Data Science, School of Computer Science, Queensland University of Technology, Brisbane, Queensland 4000, Australia\\ \email{kimthanh.vu@hdr.qut.edu.au, \{r.nayak,thirunavukarasu.balas\}@qut.edu.au} \inst{1}\\}
\maketitle              
\begin{abstract}
Modern businesses are increasingly challenged by the time and expense required to generate and assess high-quality content. Human writers face time constraints, and extrinsic evaluations can be costly. While Large Language Models (LLMs) offer potential in content creation, concerns about the quality of AI-generated content persist. Traditional evaluation methods, like human surveys, further add operational costs, highlighting the need for efficient, automated solutions. This research introduces Generative Agents as a means to tackle these challenges. These agents can rapidly and cost-effectively evaluate AI-generated content, simulating human judgment by rating aspects such as coherence, interestingness, clarity, fairness, and relevance. By incorporating these agents, businesses can streamline content generation and ensure consistent, high-quality output while minimizing reliance on costly human evaluations. The study provides critical insights into enhancing LLMs for producing business-aligned, high-quality content, offering significant advancements in automated content generation and evaluation.

\end{abstract}

\section{Introduction}
\label{introduction}
Natural Language Generation (NLG) has rapidly emerged as a transformative technology that enhances the efficiency of content creation. Owing to its ability to generate human-like content, NLG has attracted considerable interest for its application in various industries such as healthcare \cite{bhirud2019literature}, finance, and marketing \cite{skrodelis_latest_2023}. Advances in NLG models \cite{brown2020language} have facilitated improvements in text generation that minimize the need for human editing in areas like storytelling \cite{alabdulkarim2021automatic} and advertising \cite{kabaso2020sell}.

However, AI-generated text faces significant challenges in achieving improved human-like writing quality due to (1) the absence of powerful evaluation metrics and (2) the lack of standard referenced-based benchmark datasets \cite{guan2021openmeva}. A traditional approach of NLG evaluation involves comparing correlation metrics with human judgement, but this approach has shown to be ineffective \cite{guan2020union}. Additionally, popular NLG metrics such as ROUGE \cite{lin2004rouge} and BLEU \cite{papineni2002bleu} struggle to represent human-level cognition in a given text. This stems from the lack of well-annotated datasets in this field, as the cost of human annotation is high and can introduce bias. Moreover, there is no consensus on the human evaluation criteria, with various metrics proposed, such as coherence, relevance, and overall quality on a Likert scale \cite{chhun2022human}. These challenges collectively demand a more effective framework that can replace the human effort in data annotation while maintaining evaluation reliability. 

One of the prevalent questions regarding the analysis of conventional evaluation metrics is:  \textit{"How do these metrics measure the goodness of the content as humans perceive it?".} In response to this question, this paper proposes AgentEval,  a framework incorporating Generative Agents \cite{park2023generative} with chain-of-thoughts (CoT) \cite{wei2022chain} that evaluates the quality of generated texts in multi-step prompting. 
We aim to capture fairness in the content evaluation 
by assigning human personalities during agent initialization. 

Our research demonstrates that these LLM-driven agents can effectively simulate human judgments, laying the groundwork for removing human involvement from data annotation. We assert that this framework can address both reference-based and reference-free metrics by allowing the agent to articulate their understanding of the texts, which can then be used as annotations. Our agents are also asked to evaluate content as a human by providing ratings on dimensions such as coherence, interestingness, clarity, fairness and relevance 
\cite{chhun2022human}

In summary, our main contributions to this paper are: 
\begin{itemize}
\item A LLM-based framework, \textbf{\textit{AgentEval}}, mimics the human evaluation process to assess generated texts, demonstrating enhanced reliability and cost-effectiveness in data annotation.
\item \textbf{reference-free} \textbf{metrics} that integrates various cognitive and psychological aspects humans experience while reading content, which the agents can replicate and quantify. These metrics aim to be more reliable and stable through CoT and well-defined evaluation criteria than conventional approaches.
\item A detailed \textbf{\textit{Evaluation Criteria}} for content rating constituted by the opinions of LLM-driven agents and previous social studies.
\end{itemize}

\section{Related Works}
\label{Related}
\subsection{Computer-based Evaluation Metrics}
Various computer-based metrics have been developed for evaluating language generation, typically categorized based on their reliance on human-generated reference texts. Reference-based metrics such as BLEU \cite{papineni2002bleu} and ROUGE \cite{lin2004rouge} primarily measure the similarity between generated texts and predefined references using word-overlap or word embeddings, exemplified by BERTScore \cite{bert-score}. However, these metrics are inappropriate for NLG tasks where the generated texts are creative and have no prior references. To mitigate this, reference-free metrics like perplexity or UNION \cite{guan2020union} evaluate content independently of any references, focusing instead on intrinsic text properties. These metrics assess weather generated texts could plausibly be human-like based on probabilistic computation and statistical inference. However, both reference-free and reference-based metrics often struggle to fully align with human judgment, particularly in creative writing tasks \cite{guan2021openmeva}. G-Eval \cite{liu2023g} and SummaEval \cite{alhussain_automatic_2022} made a significant effort to align human judgement by introducing frameworks that utilize CoT for generating human-like ratings such as coherence. However, these frameworks exhibit biases, and the selection of human-like aspects for ratings remains vaguely defined. Another LLM-driven framework, OpenMEVA \cite{guan2021openmeva}, enhances NLG evaluation by focusing on sentence-level structure and consistency using perturbed inputs. However, its scalability is limited by heavy reliance on human-generated references, which constrains its broader applicability across diverse NLG tasks. Finally, Wang et al. (2023) proposed to directly chat with ChatGPT for rating within a single prompt \cite{wang2023chatgpt}. This idea is limited as the results tend to be biased due to the lack of defined rating criteria.

\subsection{Human-based Evaluation Metrics}
Human evaluation remains the gold standard for assessing NLG due to its ability to capture nuanced aspects of language that computer-based evaluations often overlook. Standard evaluation criteria include fluency, coherence, creativity, faithfulness, fidelity, grammar, overall quality, relevance, narrative flow, emotional resonance, and content quality \cite{nenkova2005automatic}. However, the application of these criteria varies widely, with no consensus on standard protocols, often leading to evaluations that only use two or three criteria and lack specificity. Chhun et al. (2022) \cite{chhun2022human} proposed a framework named HANNA, integrating a comprehensive set of non-redundant human criteria inspired by social science studies such as relevance, coherence, empathy, surprise, engagement, and complexity into an NLG model for AI storytelling evaluation. Despite their innovative approach, the results showed that applying these criteria is still impractical due to ongoing dependence on human evaluation.
 
We propose a more comprehensive approach than G-Eval and HANNA, leveraging CoT processing to align computer-based evaluations with human judgments. This approach aims to minimize human involvement while maintaining the depth and reliability of human evaluative standards.

\section{Proposed Methodology of evaluating AI-generated content}
\begin{figure}
    \centering
\includegraphics[width=1\linewidth]{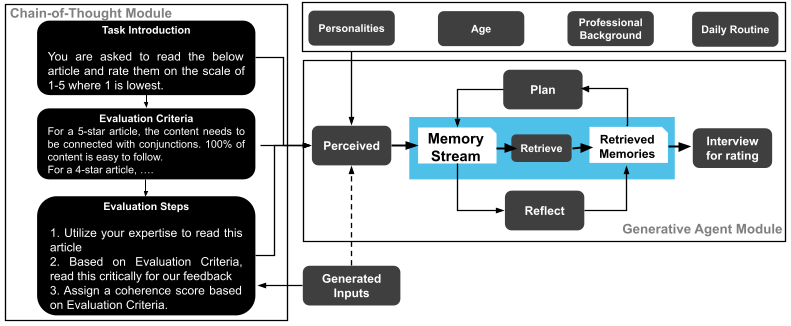}
    \caption{The framework of AgentEval includes two main components: Chain-of-Thoughts and Generative Agent. In the initial interaction with the Agent, we provide the Task introduction to prepare it for upcoming actions. We then define evaluation criteria across five dimensions: Cohenrence, Relevance, Interestingness, Fairness and Clarity. Next, we ask each agent to review our generated articles and rate them on a scale of 1 to 5.}
    \label{fig:method}
    \vspace{-2em}
\end{figure}
\label{methodology}
The proposed AgentEval framework is a comprehensive system designed to enhance the text evaluation processes by simulating human-like judgment. It comprises two main components: Chain-of-Thoughts (CoT) and Generative Agent (GA), as shown in Fig \ref{fig:method}. These modules are developed independently prior to the integration via the \textbf{\textit{Perception}} component. Below, we detail the functionality and interaction of these components. Our novel innovation lies in the integration with GA and the quantifiable \textit{Evaluation Criteria}. 
\subsection{Generative Agent Module}

This module forms the backbone of our AgentEval framework, designed to simulate human-like evaluative processes through sophisticated personalization and systematic input handling. These generative agents \cite{park2023generative} were driven by the power of LLM and take actions in an iterative process:
\vspace{-1em}
\subsubsection{Perception}
This is the connective component of two modules that takes input from two main sources: Human information and CoT commands. Each agent is personalized with unique human-like characteristics such as three personalities (\textbf{P1,P2,P3}), age, daily routine and professional background. This personal information, in our research, is provided by our participants to simulate their perspectives on content rating. Post-personalization, the agents interact with a 3-step sequential input system through CoT that will be discussed in Section \ref{COT}.
\vspace{-1em}
\subsubsection{Memory, Retrieval, Planning, Reflection Process}
Central to the agent’s cognitive process is the \textit{Memory} Stream, where all acquired \textbf{\textit{Perception}} are stored. The \textit{Retrieve }function actively pulls relevant memories from this stream, allowing the agent to engage in a \textit{Planning} phase where strategic decisions are made on approaching the evaluation based on the retrieved information. This is followed by a \textit{Reflection} phase, where the agent deliberates internally, akin to conducting an 'interview' with itself to reassess and align the content with the\textit{ Evaluation criteria}.
\vspace{-1em}
\subsubsection{Interview for Rating}
The culmination of this process is the Rating, where the agent synthesizes the insights gained from its \textit{Reflection} and \textit{Planning} to comprehensively evaluate the content. The value of the Rating is dependent on the \textit{Evaluation Criteria} and \textit{Evaluation Steps} defined in the following subsection.
\vspace{-3em}
\subsection{Chain-of-Thoughts Module}
\label{COT}
\begin{figure}
    \centering
    \includegraphics[width=0.6\linewidth]{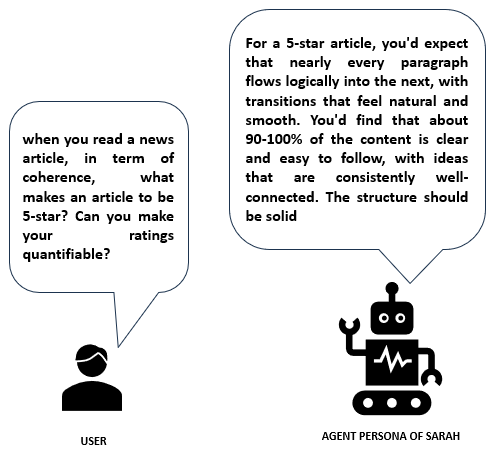}
    \caption{A sample prompt from a user to Sarah Persona (agent) about quantifying a 5-star article in terms of coherence. This prompt is repeated for lower ratings and other evaluation metrics to provide insight into each agent's thoughts on good content. We then unify all agents' responses using voting majority to develop our \textbf{Evaluation Criteria}. }
    \vspace{-2em}
    \label{fig:sarah}
\end{figure}

\vspace{-3em}
This module is a sequence of intermediate, beginning with the \textit{Task Introduction}, where the Generative Agent is informed about the upcoming evaluation task, setting expectations for the type of analysis required. This stage is crucial for establishing the context of the evaluation and ensuring the agent is primed for the specific assessment dimensions. Next, specific \textit{Evaluation Criteria}, inspired by DUC \cite{nenkova2005automatic} and psychological studies \cite{chhun2022human}, are set across five dimensions: \textbf{Coherence}, \textbf{Relevance}, \textbf{Interestingness}, \textbf{Fairness}, and \textbf{Clarity}. Our \textit{Evaluation Criteria} is further elaborated compared to prior works as we also asked agents to provide their criteria in rating content as illustrated in Fig \ref{fig:sarah}. Their responses will then be unified using a voting majority to establish a well-defined evaluation criteria included in Appendix \ref{appendixA}.
Finally, agents are tasked with \textit{Evaluation Steps} to read the generated articles and provide ratings based on these criteria, assigning scores of 1 to 5. The following are our 3-step prompt instructions for coherence rating:
\vspace{3em}
\begin{tcolorbox}[boxrule=0.5pt, colback=white, colframe=black, sharp corners]

\begin{lstlisting}[language=Python, gobble=4, basicstyle=\scriptsize\ttfamily, breaklines=true]
    1. Utilize the knowledge and expertise in your job, and read these articles carefully. You will be asked to evaluate them with the criteria provided in the next step.
    2. Evaluate the coherence of that article with the following criteria: {Evaluation criteria}
    3. Assign a coherence score on a scale of 1 to 5, where 1 is the lowest and 5 is the highest, based on the evaluation criteria.
\end{lstlisting}
\end{tcolorbox}

Each of these steps is inputted iteratively in the Perception component for an individual evaluation metric to allow agents to store the information separately in Memory effectively. This enables consistency in agent behaviours in rating AI-generated content.

\section{Experiment Design}
The objective of experiments is to answer the following question: \textit{"How effective are Generative Agents at evaluating the quality of AI-generated content compared to computer-based metrics and how they correlate with human judgement?"}.

\subsection{Data Collection}
In our experiments, we generated 30 articles using two advanced language models, Ollama 3.1 and GPT-4, each producing 15 articles with the temperature parameter set to 0.8. The content generation was guided by few-shot learning, leveraging real articles from news sources like BBC and CNN as prompts. These prompts helped the models create new 6-sentence stories with engaging titles that mimic professionally written editorial content. \textbf{We did not use existing databases for evaluation as we want to test the sensitivity of our proposed models with distinct backbones of LLMs, as discussed in Section \ref{llm}} . We then recruited 10 participants, Software Engineer and Researchers at various professional levels, as in Table \ref{tab:participant11}
to evaluate the generated texts with ratings based on five predefined aspects. Before providing their ratings, participants shared their age, job title, level of experience and three Personality traits (P1,P2,P3). The generated texts and experiment templates are detailed in our Github\footnote{https://github.com/thanh31596/AgentEval/tree/main}.
\begin{table}[]
\centering

\tiny
\begin{tabular}{|l|r|l|l|r|}
\hline
\multicolumn{1}{|c|}{\textbf{Participant}} & \multicolumn{1}{c|}{\textbf{Age}} & \multicolumn{1}{c|}{\textbf{Job}} & \multicolumn{1}{c|}{\textbf{Personality Traits}} & \multicolumn{1}{l|}{Experience} \\ \hline
Sarah & 28 & Software Developer & Curious, Analytical, Detail-oriented & 0 \\ \hline
Alex & 39 & Software Developer & Analytical, Empathetic, Assertive & 6 \\ \hline
Steve & 28 & Software Developer & Creative, Adventurous, Intuitive & 4 \\ \hline
Diana & 32 & Software Developer & Detail-oriented, Patient, Empathetic & 1 \\ \hline
Ronnie & 33 & Software Developer & Empathetic, Logical, Curious & 1 \\ \hline
Yash & 41 & Researcher & Logical, Detail-oriented, Analytical & 6 \\ \hline
Kim & 53 & Researcher & Intuitive, Creative, Emotional & 15 \\ \hline
Zoe & 28 & Researcher & Assertive, Curious, Logical & 0 \\ \hline
Charles & 34 & Researcher & Patient, Logical, Assertive & 2 \\ \hline
Mine & 27 & Researcher & Adventurous, Persuasive, Energetic & 0 \\ \hline
\end{tabular}
\caption{A list of participants were involved in our annotation process. Our assumption is that Gender and Nationality are not relevant to perspective on the content evaluation}
\vspace{-4em}
\label{tab:participant11}
\end{table}
\subsection{Baseline and Evaluation Metrics}
\label{baseline}
To establish a robust comparison for our framework, we selected several standard evaluation metrics widely used in AI-generated content assessment. We compared our AgentEval framework against the state-of-the-art frameworks:
\begin{itemize}
    \item G-Eval \footnote{https://github.com/nlpyang/geval}: Sharing the similarity with CoT in the framework, comparing with G-Eval would help to prove that using Generative Agent is a significant improvement in content evaluation for NLG tasks.
    \item Single Prompting GPT4 - 1-to-5 Assessment \cite{wang2023chatgpt}: The authors regard ChatGPT as a human evaluator. It is similar to our work in the idea of chatting with an LLM for a rating, but no evaluation criteria were mentioned in their work. 
\end{itemize} These frameworks were chosen for their prevalence in the field and their established use of LLMs in evaluating the quality of AI-generated text. Our goal was to demonstrate that \textbf{AgentEval} offers superior correlation with human judgment, providing a more nuanced and accurate assessment of content quality than these traditional frameworks.

To understand how two groups (agents and human) are well aligned, we used Analysis of Variance (ANOVA)\cite{field2024discovering} to determine which variations between group means are significant to test our null hypothesis $\Hat{H}_0$ that: "Agent is aligned with Human in judgement simulation".
The high values of $p-values$ would help to accept the null hypothesis, which proves our research goal. 

Additionally, we want to know the rating difference between our AgentEval framework and state-of-the-art frameworks. Two popular metrics for this assessment are Root Mean Squared Error (RMSE) and Mean Absolute Error (MAE). 
\begin{equation}
\text{RMSE} = \sqrt{\frac{1}{n} \sum_{i=1}^n ({humanRating}_i - {Candidate Rating})_i)^2}
\end{equation}
\begin{equation}
\text{MAE} = \frac{1}{n} \sum_{i=1}^n |{humanRating}_i - {Candidate Rating}_i|
\end{equation}
where $n$ is the total number of articles ($n$=30 in our experiment), and $Candidate Rating$ implies the rating generated by each baseline framework.

Finally, to evaluate the correlation between the ratings the Agent and Human give, we use the Pearson correlation coefficient \cite{cohen2009pearson}, which is also adopted in G-eval and 1-to-5. The Pearson correlation measures the strength and direction of the linear relationship between two variables. For each agent, the formula for calculating the Pearson correlation coefficient \( r \) is given by:
\begin{equation}
r = \frac{\sum_{i=1}^{n} (X_i - \bar{X})(Y_i - \bar{Y})}{\sqrt{\sum_{i=1}^{n} (X_i - \bar{X})^2 \sum_{i=1}^{n} (Y_i - \bar{Y})^2}}
\end{equation}

where \( X_i \) and \( Y_i \) are the individual ratings provided by the Agent and Human, respectively. The Pearson evaluation will be repeated for all ten pairs of human-agent
\section{Result Analysis}

\label{result}

In this section, we evaluate the ratings humans and intelligent agents provided based on their correlation. We conducted statistical tests to observe how likely each metric aligns with the given human judgement. Before the evaluation, we show the prompting effects on the answers of initialized agents in Fig \ref{fig:FULLFEATURES}. Future works would focus on the best prompting templates for a more improved perspective generation from the agents. 
\subsection{Impact of Human Profiles in Ratings}
\label{impact}
\begin{figure}
    \centering
\includegraphics[width=1\linewidth]{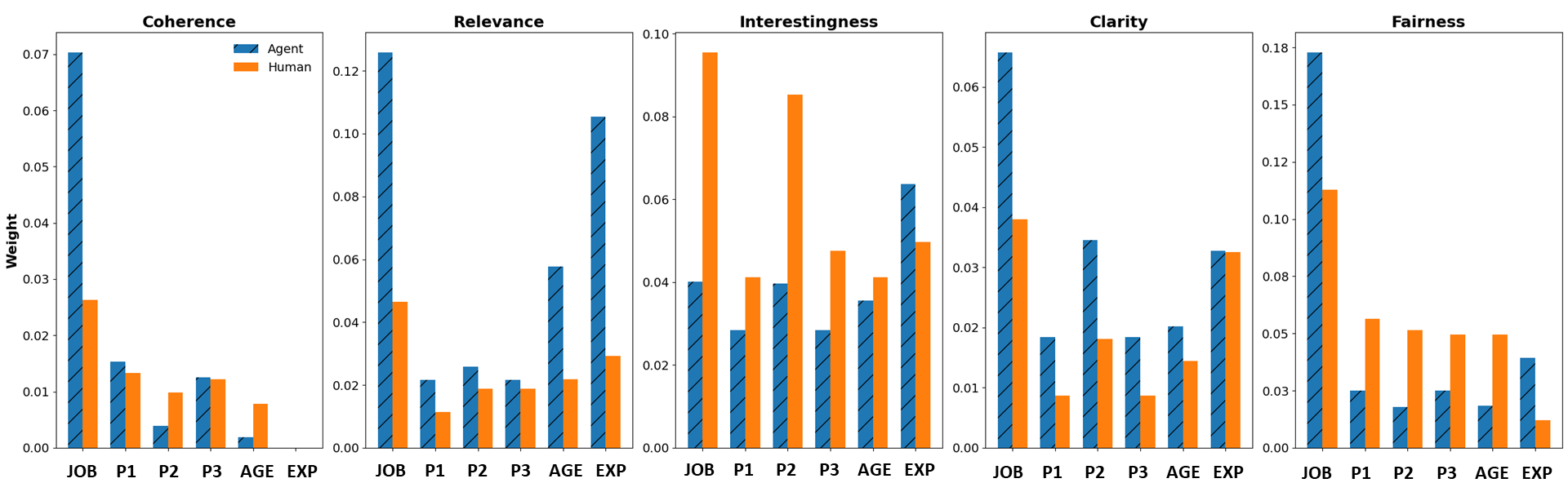}
    \caption{Feature Importance Across Rating Dimensions. The diagram illustrates the importance of various features on individual rating dimensions. In most evaluation dimensions, the agent's assessments are well-aligned with human judgments, particularly in recognizing that 'Job' is the most significant feature. However, in the 'Interestingness' dimension, the agent diverges from human judgment, placing greater importance on 'Experience' (EXP) over 'Job'}
    \label{fig:FULLFEATURES}
\end{figure}
Fig \ref{fig:FULLFEATURES} demonstrates that our agents are largely aligned with human judgements based on feature importance. We utilise Simple Linear Regression \cite{zou2003correlation} to extract the weight of each individual characteristics on our rating across five dimensions for this evaluation insight. The higher weight values imply the more important of the given characteristics to provide higher ratings on articles. In four of the five aspects (coherence, relevance, clarity, and fairness), both the agent and human evaluators consistently identify 'JOB' as the most influential feature, underscoring its critical role in determining ratings. This strong alignment suggests the \textit{Perception} component effectively stores the inputs from the users, allowing the subsequent components to simulate successfully. However, a notable divergence occurs in the 'Interestingness' aspect, where the agent assigns a higher importance to 'Experience' (EXP) than to 'JOB', contrasting with the human evaluators who still consider 'JOB' the more significant factor. This discrepancy highlights a specific area for improvement: the definition of 'Interestingness' is subjective, and our quantification is not optimized. Hence, this analysis verifies that our agents successfully mimic human judgement and it is important to receive personal information from real human evaluators to provide accurate and reliable feedback for AI-generated content

\subsection{Do personalities matter in agent initialisation for content evaluation?}
To further explore the impact of personality traits on content evaluation, we conducted a detailed analysis by weighting one-hot encoded personalities across various rating aspects. There are 14 distinct personality traits informed by 30 human participants. Table \ref{tab:weight} highlight the position-wise alignment between the agent and human evaluators varied across the personality traits on 5 rating dimensions. For coherence, relevance, and clarity, there was one correct alignment in the ranked positions of personality importance. Interestingly, in the 'Interestingness' dimension, the agent achieved three correct position-wise alignments with human evaluators, indicating a stronger agreement. This finding contrasts with the low correlation displayed in Fig \ref{fig:FULLFEATURES}. Hence, this insight affirms the significant influence of 'JOB' on rating decisions, surpassing the impact of personality traits. One noticeable observation from Table \ref{tab:weight} is that several traits is that several traits share identical weight values, suggesting the potential benefit of clustering them to enhance simplicity and efficiency in future analyses. Indeed, this finding, coupled with the analysis from Section \ref{impact}, highlights the importance of gathering personal information from users to increase the reliability of our feedback.

\begin{table}[]
\centering
\tiny
\begin{tabular}{|cccc|cccc}
\hline
\multicolumn{2}{|c|}{\textbf{Human}} & \multicolumn{2}{c|}{\textbf{Agent}} & \multicolumn{2}{c|}{\textbf{Human}} & \multicolumn{2}{c|}{\textbf{Agent}} \\ \hline
\multicolumn{4}{|c|}{\textbf{Coherence}} & \multicolumn{4}{c|}{\textbf{Relevance}} \\ \hline
\multicolumn{1}{|c|}{\textit{Persuasive}} & \multicolumn{1}{c|}{0.028} & \multicolumn{1}{c|}{\textit{Persuasive}} & 0.032 & \multicolumn{1}{c|}{Patient} & \multicolumn{1}{c|}{0.042} & \multicolumn{1}{c|}{Adventurous} & \multicolumn{1}{c|}{0.073} \\ \hline
\multicolumn{1}{|c|}{Adventurous} & \multicolumn{1}{c|}{0.028} & \multicolumn{1}{c|}{Empathetic} & 0.032 & \multicolumn{1}{c|}{Analytical} & \multicolumn{1}{c|}{0.032} & \multicolumn{1}{c|}{Energetic} & \multicolumn{1}{c|}{0.073} \\ \hline
\multicolumn{1}{|c|}{Energetic} & \multicolumn{1}{c|}{0.028} & \multicolumn{1}{c|}{Patient} & 0.023 & \multicolumn{1}{c|}{Curious} & \multicolumn{1}{c|}{0.032} & \multicolumn{1}{c|}{Persuasive} & \multicolumn{1}{c|}{0.073} \\ \hline
\multicolumn{1}{|c|}{Detail-oriented} & \multicolumn{1}{c|}{0.025} & \multicolumn{1}{c|}{Analytical} & 0.023 & \multicolumn{1}{c|}{\textit{Detail-oriented}} & \multicolumn{1}{c|}{0.032} & \multicolumn{1}{c|}{\textit{Detail-oriented}} & \multicolumn{1}{c|}{0.057} \\ \hline
\multicolumn{1}{|c|}{Curious} & \multicolumn{1}{c|}{0.025} & \multicolumn{1}{c|}{Detail-oriented} & 0.023 & \multicolumn{1}{c|}{Logical} & \multicolumn{1}{c|}{0.031} & \multicolumn{1}{c|}{Curious} & \multicolumn{1}{c|}{0.039} \\ \hline
\multicolumn{4}{|c|}{\textbf{Clarity}} & \multicolumn{4}{c|}{\textbf{Interestingness}} \\ \hline
\multicolumn{1}{|c|}{Patient} & \multicolumn{1}{c|}{0.064} & \multicolumn{1}{c|}{Empathetic} & 0.050 & \multicolumn{1}{c|}{\textit{Logical}} & \multicolumn{1}{c|}{0.174} & \multicolumn{1}{c|}{\textit{Logical}} & \multicolumn{1}{c|}{0.116} \\ \hline
\multicolumn{1}{|c|}{Detail-oriented} & \multicolumn{1}{c|}{0.061} & \multicolumn{1}{c|}{Curious} & 0.050 & \multicolumn{1}{c|}{\textit{Assertive}} & \multicolumn{1}{c|}{0.155} & \multicolumn{1}{c|}{\textit{Assertive}} & \multicolumn{1}{c|}{0.077} \\ \hline
\multicolumn{1}{|c|}{Curious} & \multicolumn{1}{c|}{0.061} & \multicolumn{1}{c|}{Logical} & 0.042 & \multicolumn{1}{c|}{Patient} & \multicolumn{1}{c|}{0.135} & \multicolumn{1}{c|}{Curious} & \multicolumn{1}{c|}{0.077} \\ \hline
\multicolumn{1}{|c|}{Analytical} & \multicolumn{1}{c|}{0.061} & \multicolumn{1}{c|}{Assertive} & 0.039 & \multicolumn{1}{c|}{\textit{Adventurous}} & \multicolumn{1}{c|}{0.089} & \multicolumn{1}{c|}{\textit{Adventurous}} & \multicolumn{1}{c|}{0.077} \\ \hline
\multicolumn{1}{|c|}{\textit{Empathetic}} & \multicolumn{1}{c|}{0.054} & \multicolumn{1}{c|}{\textit{Empathetic}} & 0.039 & \multicolumn{1}{c|}{Analytical} & \multicolumn{1}{c|}{0.089} & \multicolumn{1}{c|}{Patient} & \multicolumn{1}{c|}{0.068} \\ \hline
\multicolumn{4}{|c|}{\textbf{Fairness}} &  &  &  &  \\ \cline{1-4}
\multicolumn{1}{|c|}{Empathetic} & \multicolumn{1}{c|}{0.096} & \multicolumn{1}{c|}{Patient} & 0.123 &  &  &  &  \\ \cline{1-4}
\multicolumn{1}{|c|}{Analytical} & \multicolumn{1}{c|}{0.096} & \multicolumn{1}{c|}{Curious} & 0.117 &  &  &  &  \\ \cline{1-4}
\multicolumn{1}{|c|}{\textit{Detail-oriented}} & \multicolumn{1}{c|}{0.078} & \multicolumn{1}{c|}{\textit{Detail-oriented}} & 0.117 &  &  &  &  \\ \cline{1-4}
\multicolumn{1}{|c|}{Patient} & \multicolumn{1}{c|}{0.078} & \multicolumn{1}{c|}{Analytical} & 0.105 &  &  &  &  \\ \cline{1-4}
\multicolumn{1}{|c|}{\textit{Empathetic}} & \multicolumn{1}{c|}{0.078} & \multicolumn{1}{c|}{\textit{Empathetic}} & 0.105 &  &  &  &  \\ \cline{1-4}
\end{tabular}

\caption{Detailed personalities's on the rating. We do found that in term of initilization, types of personalities would impact greatly on content evaluation.} 
\vspace{-6em}
\label{tab:weight}
\end{table}

\subsection{Alignment with Human Judgement}
\label{llm}
\begin{figure}[H]
    \centering
    \includegraphics[width=1\linewidth]{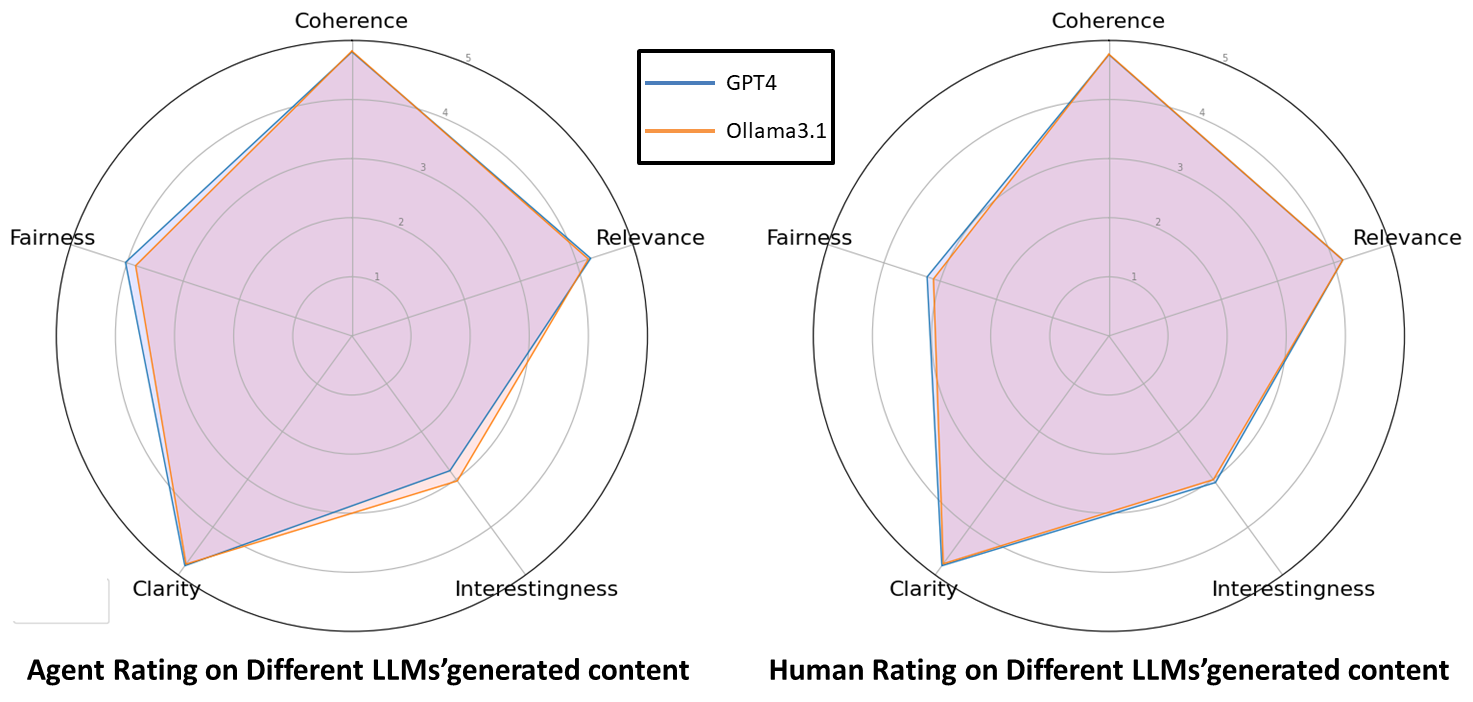}
    \caption{Average Rating on articles generated by 2 Large Language Models: GPT4 and Ollama3.1. There have not been significant differences, but while Ollama3.1 seems more interesting, GPT4 seems more fair in content writing.}
    \label{fig:radar}
    \vspace{-2em}
\end{figure}

To assess the reliability of agents' ability to mimic human judgment, we analyzed their rating distribution across various dimensions for 30 articles generated by two different LLMs: GPT-4 and Ollama 3.1. As illustrated in Fig \ref{fig:radar}, both agents and humans showed similar sensitivity in their ratings, with a consensus that articles from GPT-4 were fairer but less interesting than those from Ollama 3.1. The only notable variation between the two groups was observed in the assessment of fairness, suggesting that while agents effectively simulate human judgment in many aspects, there is room for improvement in how fairness is evaluated. 

To articulate our observations, we conducted ANOVA tests to verify the alignment between agent and human ratings. Our null hypothesis is that "Agent is aligned with human judgement". The results, presented in Table \ref{tab:anova_results}, revealed mixed levels of alignment across different metrics. Coherence, Relevance, Interestingness, and Clarity demonstrated the acceptance of the null hypothesis, evidenced by high p-values. In contrast, Fairness showed no alignment, echoing the findings from Fig \ref{fig:radar} and indicating a need for refining the evaluation criteria in this dimension. Overall, our analysis confirms that regardless of the LLM used, agent ratings align closely with human judgments, highlighting the robustness of agent evaluations in mimicking human assessment processes.

\begin{table}[ht]
\centering

\begin{tabular}{@{}lccccc@{}}
\toprule
Metrics (Errors)      &  Sum of Squares & Mean Square & F-Value & p-Value \\ \midrule
Coherence       &  0.375 (99.7) & 0.375 (0.167) & 2.248 & 0.134 \\
Relevance     &  0.667 (344.9) & 0.667 (0.577) & 1.156 & 0.283 \\
Interestingness &  1.815 (1204.9) & 1.815 (2.015) & 0.901 & 0.343 \\
Clarity     &  0.002 (100.1) & 0.002 (0.167) & 0.010 & 0.921 \\
Fairness       &  86.640 (761.4) & 86.640 (1.273) & 68.044 & \textless 0.00001 \\
Average        &  3.557 (103.9) & 3.557 (0.174) & 20.463 & 0.000007 \\
\bottomrule
\end{tabular}
\caption{ANOVA Results for Agent and Human Ratings}
\vspace{-3em}
\label{tab:anova_results}
\end{table}
\vspace{-1em}
\subsection{Is AgentEval more correlated with Human Judgement than state-of-the-art frameworks?}
The comparative analysis of AgentEval against other leading frameworks like G-Eval and 1-to-5, as presented in Table \ref{tab:rmse}, distinctly showcases AgentEval's superior alignment with human judgments across various attributes. 
For crucial attributes such as Coherence, Clarity, and Fairness, AgentEval demonstrates the lowest RMSE and MAE values, indicating a closer approximation to human evaluations than the other metrics. Specifically, AgentEval's RMSE and MAE scores in Coherence are significantly lower than those for G-Eval and 1-to-5, suggesting a more precise evaluation capability. Similarly, in attributes like Relevance and Interestingness, although 1-to-5 occasionally shows lower errors than G-Eval, AgentEval consistently outperforms both, as evidenced by the markedly lower RMSE and MAE values.
\vspace{-2em}
\begin{table}[ht]
\centering

\begin{tabular}{@{}lllll@{}}
\toprule
Attribute & Metric & AgentEval & G-Eval & 1-to-5 \\ \midrule
Coherence & RMSE & \textbf{0.166} & 1.093 & 1.296 \\
          & MAE  & \textbf{0.137} & 0.910 & 0.850 \\
Relevance & RMSE & \textbf{0.308} & 0.670 & 0.450 \\
          & MAE  & \textbf{0.240} & 0.540 & 0.390 \\
Interestingness & RMSE & \textbf{0.591} & 1.505 & 1.630 \\
                & MAE  & \textbf{0.497} & 1.397 & 1.530 \\
Clarity   & RMSE & \textbf{0.201} & 1.233 & 1.320 \\
          & MAE  & \textbf{0.163} & 1.047 & 1.120 \\
Fairness  & RMSE & \textbf{0.919} & 1.489 & 1.510 \\
          & MAE  & \textbf{0.800} & 1.390 & 1.410 \\
\bottomrule

\end{tabular}
\caption{Comparison of RMSE and MAE between AgentEval and other frameworks}
\vspace{-4em}
\label{tab:rmse}
\end{table}

A more in-depth analysis of human alignment using Pearson correlation, as shown in Table \ref{tab:pearson}, reveals that AgentEval is notably more effective in aligning with human judgment across all evaluation metrics, outperforming both G-eval and 1-to-5 approaches. Notably, in metrics where human alignment is critical, such as 'Interestingness' and 'Clarity,' AgentEval's performance far outstripped that of G-eval and 1-to-5, which showed minimal correlation, particularly in 'Interestingness' (0.01 for G-eval) and 'Clarity' (0.03 for 1-to-5). The 95\% confidence interval allows our findings to appear more robust to other frameworks regardless of which agents were being used. These results highlight the significance of AgentEval's approach, suggesting that its use of generative agents with well-defined \textbf{Evaluation Criteria }offers a more reliable and nuanced alignment with human assessments.

    
    
    
    

\section{Discussion and Conclusion}
This paper addressed the problems of reliability by introducing a novel framework, AgentEval, that utilizes combinations of Generative Agents and Chain-of-though scheme, offering a more human-aligned content assessment. Our methodology improves upon traditional referenced-free frameworks by systematically combining human-like agent profiles and structured evaluation criteria. It implies a potentially cost-effective solution for automating content evaluation that removes humans from the loop.
\vspace{-1em}
\begin{table}[H]
\centering
\begin{tabular}{|l|c|r|
>{\columncolor[HTML]{FFFFFF}}r |}
\hline
\multicolumn{1}{|c|}{\textbf{Metric}} & \textbf{AgentEval} & \multicolumn{1}{c|}{\textbf{G-eval}} & \multicolumn{1}{c|}{\cellcolor[HTML]{FFFFFF}{\color[HTML]{383838} \textbf{1-to-5}}} \\ \hline
Coherence & \textbf{0.38±0.08} & 0.32 & {\color[HTML]{383838} 0.05} \\ \hline
Relevance & \textbf{0.26±0.07} & 0.08 & {\color[HTML]{383838} 0.12} \\ \hline
Interestingness & \textbf{0.21±0.08} & 0.01 & {\color[HTML]{383838} 0.11} \\ \hline
Clarity & \textbf{0.31±0.06} & 0.05 & {\color[HTML]{383838} 0.03} \\ \hline
Fairness & \textbf{0.19±0.07} & 0.19 & {\color[HTML]{383838} 0.22} \\ \hline
\end{tabular}
\caption{Pearson Correlation among technique. We aggregated the correlation results from each agent and use 95\% confidence interval. Since G-eval adn 1-to-5 do not have multiple personalities, we compare them with the mean ratings of human evaluator. }
\vspace{-4em}
\label{tab:pearson}
\end{table}
The empirical analysis presented in Section \ref{result} validates our initial hypothesis that AgentEval exhibits a superior ability to align with human judgement across multiple content evaluation metrics using ANOVA statistical testing. The distribution of ratings across different LLMs also indicates the consistency in providing ratings in our proposed framework AgentEval. We highlighted that the profiling stage is important as it heavily impacts the manner of rating from our agents, and it should be further studied in future works in terms of clustering. Compared with other LLM-driven frameworks, AgentEval consistently outperforms in correlation and estimation error values. 

These results have significant implications for the development of automated evaluation tools. They suggest that AgentEval can be a more reliable alternative to traditional methods, particularly in industries that rely heavily on the quality assessment of generated content, such as media, academia, and online content platforms. This study recommends the broader adoption and integration of AgentEval into these industries, emphasizing the importance of continuous refinement and validation to ensure its adaptability and accuracy remain high across different contexts and content types. 

Future research should focus on extending the profiling characteristics and evaluation criteria quantification, particularly in 'Interestingness', to exploit the potential of this framework. Additionally, we want to seek a comprehensive integration approach that can fuse these evaluation dimensions into one singular value for improved interpretability and feasibility. Finally, JOB is proven to be the most influential feature on rating behaviours, yet we only collected two jobs from 10 people: Researchers and Software Engineers. Therefore, our future attempts will expand the survey to a wide variety of occupations with wider population of participants.
By addressing these areas, subsequent studies can build on the foundation laid by this research, contributing to the advancement of automated systems that are capable of delivering human-like content evaluations efficiently and accurately. 
\bibliography{reference}

\appendix
\label{appendix A}
\section{Sample of Evaluation Criteria}



\begin{table}[]
\resizebox{\columnwidth}{!}{%
\begin{tabular}{|c|l|l|l|l|l|}
\hline
\textbf{Score} & \multicolumn{1}{c|}{\textbf{Coherence}} & \multicolumn{1}{c|}{\textbf{Interestingness}} & \multicolumn{1}{c|}{\textbf{Clarity}} & \multicolumn{1}{c|}{\textbf{Relevance}} & \multicolumn{1}{c|}{\textbf{Fairness}} \\ \hline
5 & \begin{tabular}[c]{@{}l@{}}Logical progression with \\ no more than 1 minor disruption.\end{tabular} & \begin{tabular}[c]{@{}l@{}}Highly engaging; \\ with 4-5 engaging points.\end{tabular} & \begin{tabular}[c]{@{}l@{}}Clear throughout; fewer than \\ 2 unclear sections.\end{tabular} & 90-100\% relevant to the title & \begin{tabular}[c]{@{}l@{}}No bias detected; the article \\ fairly represents all relevant \\ perspectives, uses neutral \\ language, and provides a \\ balanced view of facts.\end{tabular} \\ \hline
4 & \begin{tabular}[c]{@{}l@{}}Logical flow with \\ 2-3 minor disruptions.\end{tabular} & \begin{tabular}[c]{@{}l@{}}Engaging \\ with 3-4 interesting points\end{tabular} & \begin{tabular}[c]{@{}l@{}}Mostly clear with \\ 2-3 minor unclear sections.\end{tabular} & 75-89\% relevant to the title & \begin{tabular}[c]{@{}l@{}}Mild bias detected; the article \\ may slightly favor one side, but \\ includes multiple perspectives \\ and generally neutral language.\end{tabular} \\ \hline
3 & \begin{tabular}[c]{@{}l@{}}Logical structure with \\ 3-4 disruptions present.\end{tabular} & \begin{tabular}[c]{@{}l@{}}Moderately engaging \\ with 2-3 interesting points.\end{tabular} & \begin{tabular}[c]{@{}l@{}}Moderately clear \\ with 3-4 unclear sections.\end{tabular} & 50-74\% relevant to the title. & \begin{tabular}[c]{@{}l@{}}Noticeable bias; the article \\ leans toward one viewpoint, \\ uses slightly judgmental \\ language, or lacks adequate \\ representation of all sides.\end{tabular} \\ \hline
2 & \begin{tabular}[c]{@{}l@{}}Poor logical flow with \\ 4-5 disruptions present.\end{tabular} & \begin{tabular}[c]{@{}l@{}}Slightly engaging \\ with 1-2 interesting points.\end{tabular} & \begin{tabular}[c]{@{}l@{}}Somewhat unclear with \\ 5-6 ambiguous sections.\end{tabular} & 25-49\% relevant to the title & \begin{tabular}[c]{@{}l@{}}Significant bias; the article \\ focuses on one perspective, \\ often uses emotive or biased \\ language, and omits critical \\ opposing views.\end{tabular} \\ \hline
1 & \begin{tabular}[c]{@{}l@{}}No logical progression \\ 5 or more major disruptions.\end{tabular} & \begin{tabular}[c]{@{}l@{}}Not engaging; \\ 0-1 interesting points.\end{tabular} & \begin{tabular}[c]{@{}l@{}}Very unclear with more \\ than 6 confusing sections.\end{tabular} & \textless 25\% relevant to the title & \begin{tabular}[c]{@{}l@{}}Highly biased; the article is \\ entirely one-sided, uses \\ manipulative language, and \\ excludes relevant context \\ or alternative viewpoints.\end{tabular} \\ \hline
\end{tabular}%
}
\caption{A sample of Quantified Evaluation Criteria. A detailed experiment documentation is attached in our Github}
\label{appendixA}
\end{table}

\end{document}